\documentclass[10pt,twocolumn,letterpaper]{article}

\usepackage[pagenumbers]{cvpr} 

\definecolor{cvprblue}{rgb}{0.21,0.49,0.74}
\usepackage[pagebackref,breaklinks,colorlinks,allcolors=cvprblue]{hyperref}

\def\paperID{*****} 
\def\confName{CVPR}
\def\confYear{2026}

\title{
Evidence-Based Actor–Verifier Reasoning for Echocardiographic Agents
}

\newcommand{\authorcell}[2]{%
  \parbox[t][#1][t]{0.31\linewidth}{\centering #2}%
}

\author{
\centering
\begin{tabular}{@{}ccc@{}}
\authorcell{6em}{Peng Huang\thanks{Visiting researcher at University at Albany, SUNY.}\\\
University at Albany, SUNY\\
Southwest Jiaotong University\\
{\tt\small phuang@albany.edu\\ huangpeng@my.swjtu.edu.cn }}
&
\authorcell{6em}{Yiming Wang\\
University at Albany, SUNY\\
{\tt\small ywang98@albany.edu}}
&
\authorcell{4.5em}{Yineng Chen\\
University at Albany, SUNY\\
{\tt\small ychen77@albany.edu}}
\\
\authorcell{4.5em}{Liangqiao Gui\\
University at Albany, SUNY\\
{\tt\small lgui@albany.edu}}
&
\authorcell{6em}{Hui Guo\\
University at Albany, SUNY\\
{\tt\small hguo@albany.edu}}
&
\authorcell{4.5em}{Bo Peng\thanks{Corresponding authors.}
\\
Southwest Jiaotong University\\
{\tt\small bpeng@swjtu.edu.cn}}
\\ [-1em]
\authorcell{4.5em}{Shu Hu\\
Purdue University\\
{\tt\small hu968@purdue.edu}}
&
\authorcell{4.5em}{Xi Wu\\
Chengdu University of\\
Information Technology\\
{\tt\small wuxi@cuit.edu.cn}}
&
\authorcell{4.5em}{Tsao Connie\\
Harvard Medical School\\
{\tt\small ctsao1@bidmc.harvard.edu}}
\\
\authorcell{5em}{Hongtu Zhu\\
University of North Carolina at\\
Chapel Hill\\
{\tt\small htzhu@email.unc.edu}}
&
\authorcell{5em}{Balakrishnan Prabhakaran\\
University at Albany, SUNY\\
{\tt\small bprabhakaran@albany.edu}}
&
\authorcell{5em}{Xin Wang\footnotemark[2]\\
University at Albany, SUNY\\
{\tt\small xwang56@albany.edu}}
\end{tabular}
}

\begin{document}
\maketitle
\begin{abstract}
Echocardiography plays an important role in the screening and diagnosis of cardiovascular diseases. However, automated intelligent analysis of echocardiographic data remains challenging due to complex cardiac dynamics and strong view heterogeneity. In recent years, visual language models (VLM) have opened a new avenue for building ultrasound understanding systems for clinical decision support. Nevertheless, most existing methods formulate this task as a direct mapping from video and question to answer, making them vulnerable to template shortcuts and spurious explanations. To address these issues, we propose EchoTrust, an evidence-driven Actor-Verifier framework for trustworthy reasoning in echocardiography VLM-based agents. EchoTrust produces a structured intermediate representation that is subsequently analyzed by distinct roles, enabling more reliable and interpretable decision-making for high-stakes clinical applications.
\end{abstract}    
\section{Introduction}
\label{sec:intro}

Echocardiography is one of the most widely used imaging modalities in clinical cardiovascular diagnosis and management, owing to its advantages of being real-time, non-invasive, cost-effective, and capable of simultaneously assessing cardiac structure and function~\cite{tseng2022future}. These strengths make it indispensable in modern cardiovascular care, yet they also introduce distinctive challenges for algorithmic modeling. Unlike natural images or static medical scans, echocardiographic input typically consists of noisy video sequences with blurred boundaries and low signal-to-noise ratios. Many critical anatomical structures and pathological signs are visible only under specific acoustic windows and standard views. Moreover, clinically meaningful interpretation often depends not on isolated texture cues from a single frame, but on dynamic motion patterns across the cardiac cycle, particularly during systole and diastole. These characteristics make echocardiography a demanding testbed for medical visual language models (VLMs), where reliable understanding requires not only multimodal alignment but also temporally grounded and clinically coherent reasoning~\cite{chi2025echollm,li2025echovlm, hu2025improving, zheng2024contextual, lin2024robust}.

With the rapid progress of vision-language models and multimodal large models, medical VLMs have emerged as a promising paradigm for connecting image and video understanding with clinically meaningful language-based interaction~\cite{bogaert2025bridging,rao2025multimodal}. Beyond conventional recognition or classification, these systems are increasingly expected to support more general agentic behaviors, such as organizing visual evidence, performing multi-step inference, generating interpretable justifications, and deciding when to defer under uncertainty~\cite{zhang2025improve, tsai2024uu, tsai2024uu2, zhu2024cgd}. In this broader context, question answering is better viewed as one application interface for evaluating whether a medical VLM agent can reason over visual evidence in a clinically meaningful way, rather than as the sole target of modeling itself.


Despite these advances, existing medical VLMs and multimodal agents still suffer from three major limitations in echocardiography. First, many methods reduce the task to a one-step mapping from video and language input to final output, overlooking key intermediate states such as structural visibility, view consistency, and temporal evidence sufficiency. As a result, the reasoning process often fails to align with clinical decision logic. Second, because benchmark tasks frequently contain templated language forms, closed-set candidate options, and relatively stable output distributions, models may exploit textual shortcuts or dataset priors rather than extracting truly verifiable evidence from the video itself. Third, most existing methods implicitly assume that the model should always provide an answer, while paying insufficient attention to the clinically crucial question of when the model should refrain from making a prediction. Consequently, even under insufficient evidence, mismatched views, or high uncertainty, the model may still produce uncalibrated high-confidence outputs~\cite{vaid2024local}.

These issues are especially critical in high-stakes medical scenarios, where the central question is not merely whether a model can produce a correct response, but whether that response is supported by appropriate, sufficient, and auditable evidence. A clinically deployable medical VLM agent should therefore possess the ability to organize explicit evidence, expose an inspectable reasoning process, and actively abstain when necessary, rather than solely optimizing final-task accuracy. 

Motivated by these considerations, we propose \textbf{EchoTrust}, an evidence-driven Actor-Verifier framework for trustworthy reasoning in echocardiography VLM-based agents. Rather than treating the task as unconstrained answer generation, EchoTrust reformulates medical multimodal reasoning as an evidence-driven selective decision process. Specifically, we first perform domain alignment of a foundation multimodal model via LoRA-based parameter-efficient adaptation, enabling the model to better capture the visual characteristics of echocardiographic videos and the semantics of clinical language interaction. During inference, an Actor actively searches for and organizes task-relevant evidence from the echocardiographic video, generates a candidate output, and produces a set of interpretable structured evidence states, including structural visibility, view compatibility, evidence items, and associated confidence scores. A Verifier then evaluates the Actor's output together with its evidence chain, performs consistency checking and reliability assessment, and makes corrections when necessary. 

Importantly, the reasoning emphasized in this work is not a text-only chain-of-thought process that produces superficially plausible verbal explanations. Instead, it is a structured decision process grounded in verifiable visual evidence. EchoTrust treats auditable intermediate variables as first-class components of reasoning, regards abstention as a primary prediction option, and elevates confidence from an auxiliary by-product to an explicitly modeled object. In doing so, it provides a more robust, transparent, and clinically aligned technical pathway for trustworthy medical video reasoning.

The main contributions of this work are summarized as follows:
\begin{enumerate}
    \item We have reformulated the reasoning process from an unconstrained answer generation task into an evidence-driven question-answering task, explicitly emphasizing the foundational roles of structural visibility, view consistency, and evidentiary sufficiency in echocardiography question-answering tasks.
    
    \item We propose EchoTrust, a trustworthy Actor-Verifier framework that produces structured evidence states and performs answer verification, correction, and abstention based on the generated evidence chain.

    \item We introduce a unified trustworthy reasoning paradigm for medical video question answering by integrating auditable intermediate evidence, abstention as a first-class prediction option, and explicit confidence modeling within a single framework, thereby improving interpretability, robustness, and clinical usability in high-risk settings.
    
\end{enumerate}

\section{Related works}\label{related_works}
\subsection{Task-Specific AI for Echocardiography}
Existing AI research in echocardiography has primarily focused on relatively closed, task-specific settings, such as view classification, cardiac chamber segmentation, ejection fraction estimation, wall motion analysis, and structured report extraction~\cite{du2025medical}. These studies have demonstrated that neural networks can recover clinically meaningful structural and functional information from echocardiographic videos, highlighting the feasibility of automated interpretation in this imaging modality~\cite{huang2024robustly,krubha2025robust}. In particular, task-specific models have achieved promising performance in standardizing view recognition, quantifying cardiac function, and supporting workflow automation in routine clinical practice~\cite{tavakoli2025generative}.
However, most of these approaches are built around predefined label spaces and task-specific pipelines, where the prediction target is known in advance and the reasoning scope is tightly constrained by the task formulation. As a result, they are not naturally suited to open-ended question answering scenarios in which the model must flexibly connect dynamic visual observations with diverse natural language queries. This limitation motivates a shift from isolated task optimization toward more general multimodal reasoning frameworks that can support broader forms of clinical interaction~\cite{hu2025rlministyler}.

\subsection{Medical Vision-Language Models}
Medical vision-language models (VLMs) have been increasingly studied in radiology, pathology, and broader multimodal clinical scenarios, demonstrating the potential of multimodal systems to connect imaging content with clinically meaningful language-based interaction~\cite{chao2026echoatlas,quinlan2026emerging}. Prior work has shown that such models can learn cross-modal correspondences between medical images and text, enabling a range of capabilities including recognition, report-related reasoning, and multimodal knowledge integration. These developments suggest that medical VLMs provide a promising foundation for building more interactive and clinically useful AI systems~\cite{song2025teacher,liu2025medchat,bansal2025robust, wang2024u, hu2024umednerf, wang2024neural}.

However, much of the existing literature remains centered on static-image settings, where model outputs are often determined primarily by localized appearance cues within a single image. In contrast, echocardiography presents a substantially more challenging scenario, as clinically meaningful interpretation often depends on reasoning over dynamic image sequences. Moreover, echocardiographic understanding is uniquely constrained by the joint effects of acoustic window, imaging plane, and temporal phase, all of which may critically determine whether a clinical query is answerable and what evidence should be considered relevant. These characteristics distinguish echocardiography from conventional medical image-language tasks and make direct transfer of existing methods insufficient~\cite{yang2025llm}.

In parallel, recent progress in general-purpose vision-language foundation models has advanced video understanding, instruction following, and multi-step reasoning, motivating the emergence of VLM-based agents as a more general paradigm for medical AI. Compared with conventional task-specific pipelines, such agents offer the potential to organize evidence, perform structured inference, generate interpretable outputs, and support selective decision-making. This makes them particularly attractive for echocardiographic analysis, where reliable interpretation requires not only multimodal perception but also clinically coherent reasoning over temporally evolving visual evidence.

Despite this promise, directly applying existing VLMs or agentic multimodal systems to medical ultrasound remains challenging. These models may still rely on superficial correlations, fail to expose clinically meaningful intermediate evidence, and produce overconfident predictions under uncertainty. Furthermore, many current approaches emphasize final-output generation while paying insufficient attention to whether the decision process is explicitly grounded, inspectable, and reliable. These limitations underscore the need for trustworthy multimodal reasoning frameworks that are explicitly designed for evidence grounding, structured verification, and selective decision-making in high-stakes medical video settings.

\section {Methodology}\label{method}
\subsection{Evidence-Based Actor--Verifier}
\begin{figure*}
    \centering
    \includegraphics[width=0.7\linewidth]{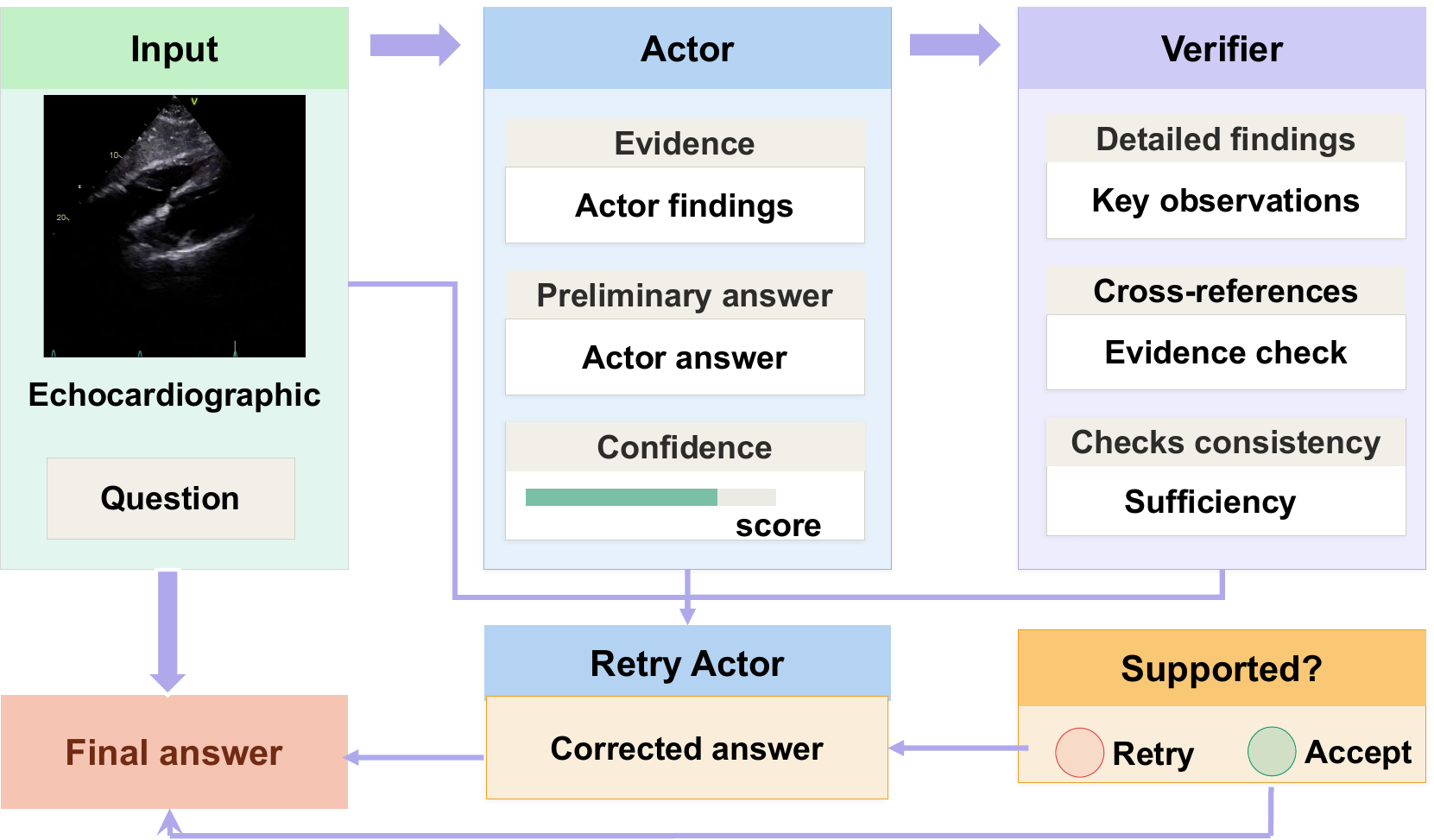}
    \caption{Overview of the proposed EchoTrust. The Actor generates evidence, a preliminary answer, and a confidence score from the input echocardiographic image and question. The Verifier then checks whether the evidence sufficiently supports the answer. If supported, the answer is accepted directly; otherwise, a Retry Actor re-infers a corrected answer guided by the Verifier's feedback. This closed-loop design enhances both answer reliability and interpretability.}
    \label{fig:overview}
\end{figure*}

We formulate trustworthy reasoning inference as a closed-loop agentic process in which prediction is not treated as a one-shot generation outcome, but as an auditable and revisable reasoning state. The central premise is that a reliable agent should not only produce a decision, but also externalize the evidential basis of that decision, expose such evidence to scrutiny, and permit corrective re-reasoning when the original inference is found to be insufficiently supported. To this end, we introduce an evidence-guided actor--verifier framework that couples answer generation with explicit evidence auditing and posterior acceptance. Fig.~\ref{fig:overview} presents an overview of the proposed EchoTrust.

Let $\mathcal{X}$ denote the input observation and $q$ the query or task instruction. The actor $\mathcal{A}$ first produces a structured reasoning state
\begin{equation}
\mathbf{s}_a = \mathcal{A}(\mathcal{X}, q),
\end{equation}
where $\mathbf{s}_a$ includes both a prediction $y_a$ and an associated evidence set $\mathbf{e}_a$, i.e.,
\begin{equation}
\mathbf{s}_a = (y_a, \mathbf{e}_a).
\end{equation}
This formulation shifts inference from an implicit black-box mapping to an explicit conclusion--evidence representation. As a result, the output of the agent becomes amenable to downstream inspection at the level of reasoning support rather than only final decisions.

Given the actor state, a verifier $\mathcal{F}$ is introduced to assess the reliability of the produced reasoning. Importantly, the verifier is not designed as an alternative predictor that competes with the actor. Instead, it serves as an evidence auditor that examines whether the current inference is adequately supported, internally coherent, and consistent with the input observation. Formally, the verifier produces
\begin{equation}
\mathbf{s}_v = \mathcal{F}(\mathcal{X}, q, \mathbf{s}_a),
\end{equation}
where $\mathbf{s}_v$ summarizes the verification outcome over the actor's reasoning state. In essence, $\mathbf{s}_v$ characterizes the degree to which the evidence in $\mathbf{s}_a$ substantiates the prediction $y_a$, while also identifying missing support, local inconsistencies, or potential conflicts. Such a design assigns the verifier the role of \emph{reasoning diagnosis} rather than decision replacement.

The actor state and verifier state are then jointly used to determine whether re-reasoning is necessary. This decision is captured by
\begin{equation}
z = \pi(\mathbf{s}_a, \mathbf{s}_v), \qquad z \in \{0,1\},
\end{equation}
where $z=1$ indicates that the current inference should be revised, and $z=0$ indicates that the original reasoning state is sufficiently trustworthy to be retained. Here, $\pi(\cdot)$ should be understood as a trust-oriented intervention policy: re-reasoning is triggered only when the current conclusion is not adequately justified by its evidential support.

When revision is required, a retry actor $\mathcal{A}_r$ performs a second-round inference conditioned on both the original input and the verifier feedback:
\begin{equation}
\mathbf{s}_r = \mathcal{A}_r(\mathcal{X}, q, \mathbf{s}_a, \mathbf{s}_v),
\end{equation}
with
\begin{equation}
\mathbf{s}_r = (y_r, \mathbf{e}_r).
\end{equation}
where $y_r$ is the revised prediction and $\mathbf{e}_r$ the corresponding evidence set. This stage is designed as controlled re-reasoning rather than direct correction. The verifier provides diagnostic guidance, but does not directly overwrite the actor's prediction. The reasoning authority therefore remains with the actor, while the verifier acts as a structured source of critique that constrains the second-pass inference.

To ensure that the revised inference indeed improves trustworthiness, the retry output is subjected to a second verification pass,
\begin{equation}
\mathbf{s}_v' = \mathcal{F}(\mathcal{X}, q, \mathbf{s}_r),
\end{equation}
followed by a posterior acceptance rule
\begin{equation}
\hat{y} =
\begin{cases}
y_r, & \text{if } \psi(\mathbf{s}_v', \mathbf{s}_v)=1,\\
y_a, & \text{otherwise}.
\end{cases}
\end{equation}
Here, $\psi(\cdot)$ determines whether the revised reasoning state achieves stronger evidential support than the original one. This posterior confirmation mechanism is critical for preventing re-reasoning from introducing unnecessary drift or replacing a plausible original conclusion with a less reliable alternative.


\subsection{Parameter-Efficient Fine-Tuning}
Initial actor, verifier, and retry actor are implemented as three role-specific modules on top of a shared frozen multimodal backbone. Let $\Theta$ denote the backbone parameters, which remain fixed throughout training, and let $\Delta\Theta_a$, $\Delta\Theta_v$, and $\Delta\Theta_r$ denote the trainable adapter parameters for the initial actor, verifier, and retry actor, respectively. Parameter-efficient low-rank adaptation is used to specialize these roles while preserving the shared representation space of the backbone.

The three modules are defined as
\begin{equation}
\mathbf{s}_a=\mathcal{A}(\mathcal{X},q;\Theta,\Delta\Theta_a),
\end{equation}
\begin{equation}
\mathbf{s}_v=\mathcal{F}(\mathcal{X},q,\mathbf{s}_a;\Theta,\Delta\Theta_v),
\end{equation}
\begin{equation}
\mathbf{s}_r=\mathcal{A}_r(\mathcal{X},q,\mathbf{s}_a,\mathbf{s}_v;\Theta,\Delta\Theta_r),
\end{equation}
where $\mathcal{X}$ denotes the input observation, $q$ denotes the query or task instruction, and $\mathbf{s}_a$, $\mathbf{s}_v$, and $\mathbf{s}_r$ denote the initial reasoning state, verification state, and retry reasoning state, respectively.

The initial actor and the retry actor are parameterized with two separate adapters, controlled by $\Delta\Theta_a$ and $\Delta\Theta_r$, respectively. The former produces an initial conclusion together with its supporting evidence from the input, while the latter performs a second-pass inference conditioned on both the initial reasoning state $\mathbf{s}_a$ and the verifier feedback $\mathbf{s}_v$. Since these two stages operate under different conditioning contexts, decoupling their adapters helps preserve the stability of initial reasoning while improving the model's ability to revise its inference under verification guidance.

The training objective of the initial actor is
\begin{equation}
\mathcal{L}_{\mathrm{actor}}
=
-\sum_{t\in\mathcal{T}_{a}}
\log p\!\left(y_t \mid y_{<t}, \mathcal{X}, q; \Theta,\Delta\Theta_a\right),
\end{equation}
where $\mathcal{T}_{a}$ denotes the set of supervised token positions in the initial actor output, and $y_t$ denotes the target token at position $t$. This objective models the generation of the structured reasoning state $\mathbf{s}_a$ from the input.

The retry actor is trained with
\begin{equation}
\mathcal{L}_{\mathrm{retry}}
=
-\sum_{t\in\mathcal{T}_{r}}
\log p\!\left(y_t \mid y_{<t}, \mathcal{X}, q, \mathbf{s}_a, \mathbf{s}_v; \Theta,\Delta\Theta_r\right),
\end{equation}
where $\mathcal{T}_{r}$ denotes the set of supervised token positions in the retry actor output. Compared with the initial actor, this objective is additionally conditioned on $\mathbf{s}_a$ and $\mathbf{s}_v$, corresponding to verification-guided re-reasoning.

The verifier is trained with
\begin{equation}
\mathcal{L}_{\mathrm{ver}}
=
-\sum_{t\in\mathcal{T}_{v}}
\log p\!\left(v_t \mid v_{<t}, \mathcal{X}, q, \mathbf{s}_a; \Theta,\Delta\Theta_v\right),
\end{equation}
where $\mathcal{T}_{v}$ denotes the set of supervised token positions in the verifier output, and $v_t$ denotes the target verifier token at position $t$. This objective learns to generate the verification state $\mathbf{s}_v$ conditioned on the input and the initial reasoning state.

Based on this design, we enable inference, verification, and correction to be learned within a unified representation space, while the Dual-Executor Adapter distinguishes initial inference from retry inference as two distinct modes of agent behavior.
\begin{figure*}[t]
    \centering
    \includegraphics[width=0.8\linewidth]{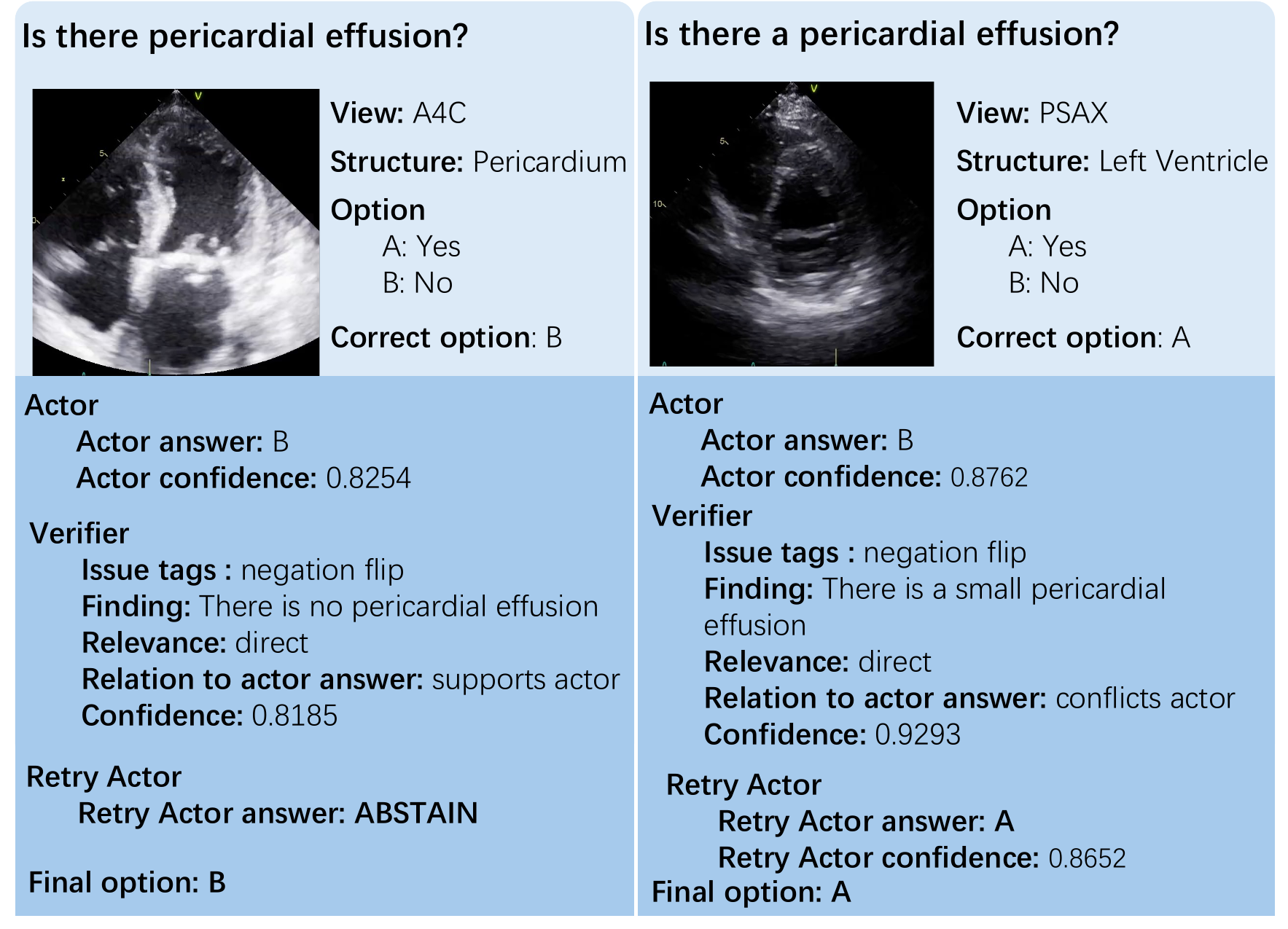}
    \caption{Case studies on binary (Yes/No) questions.}
    \label{fig:case1}
\end{figure*}

\section {Experimental results}\label{experimental}

\subsection{Dataset}
We evaluated our method on MIMICEchoQA~\cite{PhysioNet-mimic-iv,thapa2025well}, a benchmark dataset for echocardiogram-based visual question answering built upon MIMIC-IV-ECHO and matched clinical notes from MIMIC-IV-Note. Each echocardiographic study was paired with the nearest discharge summary within a ±7-day window, from which echo-specific report sections were extracted to generate clinically grounded question-answer pairs. The original DICOM studies were converted to .mp4 videos, and each video was assigned an echocardiographic view label to enable filtering of anatomically inconsistent questions. Candidate question-answer pairs were then manually reviewed by two board-certified cardiologists for clinical relevance, answer correctness, and visual answerability, resulting in a final benchmark of 622 curated video-question pairs.

Each sample in MIMICEchoQA contains a transthoracic echocardiography video, one closed-ended multiple-choice question with four candidate answers, the ground-truth answer, the queried anatomical structure, supporting report context, and metadata such as study ID, video filename, view label, and split assignment. The dataset has a strict 1:1 mapping between the question and video, covering 622 unique videos from 622 unique patients, with 48 echocardiographic views and 14 queried cardiac structures. The most frequent views include A4C, Subcostal 4-Chamber, and A3C, while commonly queried structures include the left ventricle, aortic valve, mitral valve, and pericardium. A key characteristic of MIMICEchoQA is that each question is jointly grounded in dynamic echo videos, view information, and expert-authored diagnostic reports, making it more clinically realistic than conventional medical VQA datasets centered on static images.

\subsection{Quantitative Analysis}

Table~\ref{tab:baseline} shows that directly applying general-purpose VLMs to echocardiographic question answering yields limited performance. Across different scales, the Qwen3-VL variants only achieve accuracies of 0.25--0.44, indicating that such models do not naturally possess sufficient capability for this highly specialized medical task~\cite{Qwen3-VL,Qwen-VL}. Echocardiographic reasoning depends on subtle anatomical visibility, view-specific constraints, and temporal cardiac motion, which are difficult to capture through general-domain multimodal priors alone.

Another important observation is that neither the built-in \textit{Thinking} mode nor CoT prompting provides consistent gains. In several settings, \textit{Thinking} even performs worse than \textit{Instruct}, while CoT introduces only unstable changes. This suggests that simply encouraging the model to ``think more'' does not ensure better reasoning in echocardiography. A likely reason is that general-purpose models tend to reason mainly in language space: they first form a prediction and then generate text to make that prediction appear justified, rather than grounding the answer in valid visual evidence.

\begin{figure*}[t]
    \centering
    \includegraphics[width=0.8\linewidth]{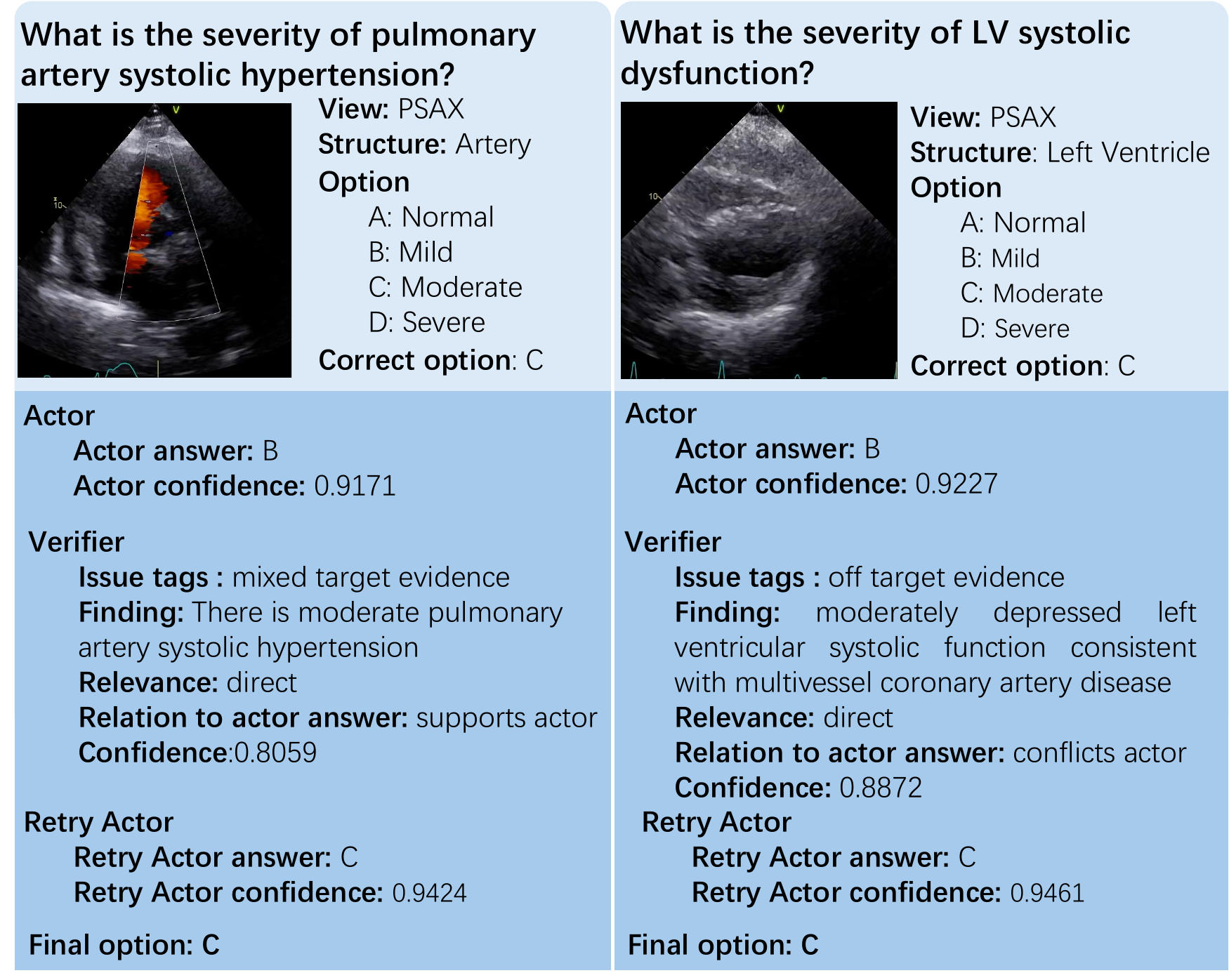}
    \caption{Case studies on four-choice severity classification questions.}
    \label{fig:case2}
\end{figure*}

In contrast, EchoTrust reformulates the task as an evidence-driven reasoning process. It first extracts and organizes structured evidence, then uses clearly differentiated roles to complete answer generation and verification. The Actor is responsible for discovering evidence and proposing an answer, whereas the Verifier evaluates the reliability of that evidence, independently searches for supporting or contradictory observations, and determines whether the conclusion is justified. If necessary, the Retry Actor reconsiders the final answer by integrating all available evidence. As shown in Table~\ref{tab:baseline}, this design improves the accuracy to 0.76, while also producing a traceable reasoning process grounded in structured and verifiable evidence, which is more suitable for clinically trustworthy deployment. Fig.~\ref{fig:case1} and ~\ref{fig:case2} provide detailed case studies.

Therefore, the advantage of EchoTrust lies in two aspects. On the one hand, it substantially improves predictive accuracy. On the other hand, it produces a traceable reasoning process in which the final answer is linked to structured evidence and explicit verification steps. Such a design is better aligned with the requirements of clinical deployment, where reliability depends not only on task performance but also on whether the decision process can be inspected, questioned, and trusted.
\begin{table}
    \centering
    \caption{Performance compared to Qwen3-VL.}
    \begin{tabular}{lc}
    \hline
        \textbf{Methods} & \textbf{Accuracy}\\
        \hline
        Qwen3-VL-2B-Instruct & 0.43\\
        Qwen3-VL-2B-Thinking & 0.25\\
        Qwen3-VL-4B-Instruct & 0.40\\
        Qwen3-VL-4B-Thinking & 0.30\\
        Qwen3-VL-8B-Instruct & 0.44\\
        Qwen3-VL-8B-Thinking & 0.34\\
        Qwen3-VL-32B-Instruct & 0.44\\
        Qwen3-VL-32B-Thinking & 0.29\\ \hline
        
        CoT: Qwen3-VL-2B-Instruct & 0.44\\
        CoT: Qwen3-VL-2B-Thinking & 0.23\\
        CoT: Qwen3-VL-4B-Instruct & 0.46\\
        CoT: Qwen3-VL-4B-Thinking & 0.26\\
        CoT: Qwen3-VL-8B-Instruct & 0.37 \\
        CoT: Qwen3-VL-8B-Thinking & 0.40\\
        CoT: Qwen3-VL-32B-Instruct & 0.37\\
        CoT: Qwen3-VL-32B-Thinking & 0.38\\ \hline
        \textbf{Ours} & \textbf{0.76}\\
        \hline
    \end{tabular}
    \label{tab:baseline}
\end{table}

\begin{table}[t]
    \centering
    \caption{The performance of the actor and the actor-verifier system.}
    \begin{tabular}{cc} \hline
    \textbf{Setting} & \textbf{Accuracy} \\ \hline
       Actor only  & 0.62 \\
       Actor + Verifier  & \textbf{0.76}\\ \hline
    \end{tabular}
    \label{tab:Component}
\end{table}

\subsection{Ablation Analysis}
We further evaluate the contribution of the verification module through a component analysis. As shown in Table~\ref{tab:Component}, using the Actor alone achieves an accuracy of 0.62, while introducing the Verifier improves the accuracy to 0.76. This result indicates that evidence verification is not merely an auxiliary post-processing step, but a key component for improving overall reliability. By explicitly checking the Actor's proposed evidence and conclusion, the Verifier helps correct a substantial portion of the errors made by single-pass reasoning.

\begin{table}[t]
\centering
\caption{Statistics of verifier intervention and final switch quality.}
\begin{tabular}{lc}
\hline
\textbf{Metric} & \textbf{Count} \\
\hline
Total samples & 100 \\
keep actor & 65 \\
retry actor & 35 \\
retry attempted & 51 \\
retry answer changed & 34 \\
switch applied & 27 \\
wrong $\rightarrow$ correct & 20 \\
correct $\rightarrow$ wrong & 6 \\
wrong $\rightarrow$ wrong & 1 \\
\hline
\end{tabular}
\label{tab:detail}
\end{table}

The detailed statistics in Table~\ref{tab:detail} further explain where this gain comes from. Among 100 samples, the system kept the original Actor answer in 65 cases and invoked retry-based reconsideration in 35 cases. Although retry was attempted 51 times, a final answer switch was applied in only 27 cases, showing that the Verifier does not revise predictions indiscriminately. More importantly, among these switched cases, 20 changed from wrong to correct, whereas only 6 changed from correct to wrong. This indicates that the Verifier is able to identify a meaningful subset of unreliable initial predictions and improve them through evidence re-examination, leading to a positive net effect on final accuracy.

Overall, these results support the role assignment in EchoTrust. The Actor is responsible for discovering initial evidence and producing candidate answers, while the Verifier provides an additional layer of reliability control by assessing evidence quality, detecting potential contradictions, and deciding whether a second-round answer is necessary. Such a design improves not only predictive performance but also the trustworthiness of the overall reasoning process.

\section {Conclusion}\label{conclusion}
In this work, we presented \textbf{EchoTrust}, an evidence-driven Actor-Verifier framework for trustworthy reasoning in echocardiography VLM-based agents. Instead of treating echocardiographic question answering as a direct mapping from video and language input to final output, EchoTrust reformulates the task as a structured and selective reasoning process centered on explicit evidence extraction, verification, and revision. Through clear role separation, the Actor is responsible for discovering task-relevant evidence and proposing an initial answer, while the Verifier evaluates evidence reliability, searches for potentially contradictory observations, and determines whether re-answering is necessary. Experimental results show that this design yields clear advantages over general-purpose VLM baselines and that verification plays a substantial role in improving final performance. More importantly, EchoTrust produces a traceable reasoning process grounded in structured and verifiable evidence, making it better aligned with the reliability requirements of high-stakes clinical applications. Nevertheless, our current framework may still revise some initially correct predictions into incorrect ones, indicating that its decision stability remains to be further improved. These findings suggest that moving from answer generation to evidence-supported decision-making is a promising direction for trustworthy medical VLM systems.

{
    \small
    \bibliographystyle{ieeenat_fullname}
    \bibliography{main}

@article{PhysioNet-mimic-iv,
  author = {Thapa, Rahul and Li, Andrew and Wu, Qingyang and He, Bryan and Sahashi, Yuki and Binder-Rodriguez, Christina and Zhang, Angela and Ouyang, David and Zou, James},
  title = {{MIMIC-IV-ECHO-Ext-MIMICEchoQA: A Benchmark Dataset for Echocardiogram-Based Visual Question Answering}},
  journal = {{PhysioNet}},
  year = {2025},
  month = oct,
  note = {Version 1.0.0},
  doi = {10.13026/rndk-4s36},
  url = {https://doi.org/10.13026/rndk-4s36}
}

@article{thapa2025well,
  title={How Well Can General Vision-Language Models Learn Medicine By Watching Public Educational Videos?},
  author={Thapa, Rahul and Li, Andrew and Wu, Qingyang and He, Bryan and Sahashi, Yuki and Binder, Christina and Zhang, Angela and Athiwaratkun, Ben and Song, Shuaiwen Leon and Ouyang, David and others},
  journal={arXiv preprint arXiv:2504.14391},
  year={2025}
}

@article{Qwen3-VL,
      title={Qwen3-VL Technical Report}, 
      author={Shuai Bai and Yuxuan Cai and Ruizhe Chen and Keqin Chen and Xionghui Chen and Zesen Cheng and Lianghao Deng and Wei Ding and Chang Gao and Chunjiang Ge and Wenbin Ge and Zhifang Guo and Qidong Huang and Jie Huang and Fei Huang and Binyuan Hui and Shutong Jiang and Zhaohai Li and Mingsheng Li and Mei Li and Kaixin Li and Zicheng Lin and Junyang Lin and Xuejing Liu and Jiawei Liu and Chenglong Liu and Yang Liu and Dayiheng Liu and Shixuan Liu and Dunjie Lu and Ruilin Luo and Chenxu Lv and Rui Men and Lingchen Meng and Xuancheng Ren and Xingzhang Ren and Sibo Song and Yuchong Sun and Jun Tang and Jianhong Tu and Jianqiang Wan and Peng Wang and Pengfei Wang and Qiuyue Wang and Yuxuan Wang and Tianbao Xie and Yiheng Xu and Haiyang Xu and Jin Xu and Zhibo Yang and Mingkun Yang and Jianxin Yang and An Yang and Bowen Yu and Fei Zhang and Hang Zhang and Xi Zhang and Bo Zheng and Humen Zhong and Jingren Zhou and Fan Zhou and Jing Zhou and Yuanzhi Zhu and Ke Zhu},
	  journal={arXiv preprint arXiv:2511.21631},
      year={2025}
}

@article{Qwen-VL,
  title={Qwen-VL: A Versatile Vision-Language Model for Understanding, Localization, Text Reading, and Beyond},
  author={Bai, Jinze and Bai, Shuai and Yang, Shusheng and Wang, Shijie and Tan, Sinan and Wang, Peng and Lin, Junyang and Zhou, Chang and Zhou, Jingren},
  journal={arXiv preprint arXiv:2308.12966},
  year={2023}
}

@inproceedings{zhang2025improve,
  title={Improve vision language model chain-of-thought reasoning},
  author={Zhang, Ruohong and Zhang, Bowen and Li, Yanghao and Zhang, Haotian and Sun, Zhiqing and Gan, Zhe and Yang, Yinfei and Pang, Ruoming and Yang, Yiming},
  booktitle={Proceedings of the 63rd Annual Meeting of the Association for Computational Linguistics (Volume 1: Long Papers)},
  pages={1631--1662},
  year={2025}
}

@article{tseng2022future,
  title={Future guidelines for artificial intelligence in echocardiography},
  author={Tseng, Andrew S and Lopez-Jimenez, Francisco and Pellikka, Patricia A},
  journal={Journal of the American Society of Echocardiography},
  volume={35},
  number={8},
  pages={878--882},
  year={2022},
  publisher={Elsevier}
}

@article{vaid2024local,
  title={Local large language models for privacy-preserving accelerated review of historic echocardiogram reports},
  author={Vaid, Akhil and Duong, Son Q and Lampert, Joshua and Kovatch, Patricia and Freeman, Robert and Argulian, Edgar and Croft, Lori and Lerakis, Stamatios and Goldman, Martin and Khera, Rohan and others},
  journal={Journal of the American Medical Informatics Association},
  volume={31},
  number={9},
  pages={2097--2102},
  year={2024},
  publisher={Oxford University Press}
}

@article{chi2025echollm,
  title={EchoLLM: extracting echocardiogram entities with light-weight, open-source large language models},
  author={Chi, Jonathan and Rouphail, Yazan and Hillis, Ethan and Ma, Ningning and Nguyen, An and Wang, Jane and Hofford, Mackenzie and Gupta, Aditi and Lyons, Patrick G and Wilcox, Adam and others},
  journal={JAMIA open},
  volume={8},
  number={4},
  pages={ooaf092},
  year={2025},
  publisher={Oxford University Press}
}

@article{li2025echovlm,
  title={EchoVLM: Measurement-Grounded Multimodal Learning for Echocardiography},
  author={Li, Yuheng and Zhang, Yue and Amadou, Abdoul Aziz and Lai, Yuxiang and Zhong, Jike and Passerini, Tiziano and Comaniciu, Dorin and Sharma, Puneet},
  journal={arXiv preprint arXiv:2512.12107},
  year={2025}
}

@inproceedings{huang2024robustly,
  title={Robustly optimized deep feature decoupling network for fatty liver diseases detection},
  author={Huang, Peng and Hu, Shu and Peng, Bo and Zhang, Jiashu and Wu, Xi and Wang, Xin},
  booktitle={International Conference on Medical Image Computing and Computer-Assisted Intervention},
  pages={68--78},
  year={2024},
  organization={Springer}
}

@inproceedings{krubha2025robust,
  title={Robust AI-generated face detection with imbalanced data},
  author={Krubha, Yamini Sri and Hou, Aryana and Vester, Braden and Walker, Web and Wang, Xin and Lin, Li and Hu, Shu},
  booktitle={2025 IEEE 8th International Conference on Multimedia Information Processing and Retrieval (MIPR)},
  pages={470--476},
  year={2025},
  organization={IEEE}
}

@article{hu2025rlministyler,
  title={RLMiniStyler: Light-weight RL Style Agent for Arbitrary Sequential Neural Style Generation},
  author={Hu, Jing and Feng, Chengming and Hu, Shu and Chang, Ming-Ching and Li, Xin and Wu, Xi and Wang, Xin},
  journal={arXiv preprint arXiv:2505.04424},
  year={2025}
}

@inproceedings{song2025teacher,
  title={Teacher Encoder-Student Decoder Denoising Guided Segmentation Network for Anomaly Detection},
  author={Song, Shixuan and Chen, Hao and Hu, Shu and Wang, Xin and Hu, Jinrong and Wu, Xi},
  booktitle={International Conference on Neural Information Processing},
  pages={238--253},
  year={2025},
  organization={Springer}
}

@inproceedings{liu2025medchat,
  title={Medchat: A multi-agent framework for multimodal diagnosis with large language models},
  author={Liu, Philip R and Bansal, Sparsh and Dinh, Jimmy and Pawar, Aditya and Satishkumar, Ramani and Desai, Shail and Gupta, Neeraj and Wang, Xin and Hu, Shu},
  booktitle={2025 IEEE 8th International Conference on Multimedia Information Processing and Retrieval (MIPR)},
  pages={456--462},
  year={2025},
  organization={IEEE}
}

@inproceedings{bansal2025robust,
  title={Robust fairness vision-language learning for medical image analysis},
  author={Bansal, Sparsh and Wu, Mingyang and Wang, Xin and Hu, Shu},
  booktitle={2025 IEEE 8th International Conference on Multimedia Information Processing and Retrieval (MIPR)},
  pages={463--469},
  year={2025},
  organization={IEEE}
}

@inproceedings{yang2025llm,
  title={Llm-medqa: Enhancing medical question answering through case studies in large language models},
  author={Yang, Hang and Chen, Hao and Guo, Hui and Chen, Yineng and Lin, Ching-Sheng and Hu, Shu and Hu, Jinrong and Wu, Xi and Wang, Xin},
  booktitle={2025 International Joint Conference on Neural Networks (IJCNN)},
  pages={1--8},
  year={2025},
  organization={IEEE}
}

@article{chao2026echoatlas,
  title={EchoAtlas: A Conversational, Multi-View Vision-Language Foundation Model for Echocardiography Interpretation and Clinical Reasoning},
  author={Chao, Chieh-Ju and Asadi, Mohammad and Li, Lavonda and Ramasamy, Gokul and Pecco, Nicolo and Wang, Yu-Chiang and Poterucha, Timothy and Arsanjani, Reza and Kane, Garvan C and Oh, Jae K and others},
  journal={medRxiv},
  pages={2026--03},
  year={2026},
  publisher={Cold Spring Harbor Laboratory Press}
}

@article{quinlan2026emerging,
  title={Emerging Utility of Multimodal Large Language Models in Cardiovascular Diagnostics},
  author={Quinlan, Anna G and Tsai, Mitchell H and Zimmerman, Joshua M},
  journal={Journal of Medical Systems},
  volume={50},
  number={1},
  pages={33},
  year={2026},
  publisher={Springer}
}

@article{du2025medical,
  title={Medical Knowledge Intervention Prompt Tuning for Medical Image Classification},
  author={Du, Ye and Yu, Nanxi and Wang, Shujun},
  journal={IEEE Transactions on Medical Imaging},
  year={2025},
  publisher={IEEE}
}

@article{bogaert2025bridging,
  title={Bridging vision and text: applications and challenges of vision-language models in urological surgery},
  author={Bogaert, Wouter and Carl, Nicolas and Kowalewski, Karl-Friedrich and Michel, Maurice Stephan and Mottrie, Alexandre and De Backer, Pieter},
  journal={European urology focus},
  volume={11},
  number={1},
  pages={18--21},
  year={2025},
  publisher={Elsevier}
}

@article{rao2025multimodal,
  title={Multimodal generative AI for medical image interpretation},
  author={Rao, Vishwanatha M and Hla, Michael and Moor, Michael and Adithan, Subathra and Kwak, Stephen and Topol, Eric J and Rajpurkar, Pranav},
  journal={Nature},
  volume={639},
  number={8056},
  pages={888--896},
  year={2025},
  publisher={Nature Publishing Group UK London}
}

@article{tavakoli2025generative,
  title={Generative AI and Foundation Models in Radiology: Applications, Opportunities, and Potential Challenges},
  author={Tavakoli, Neda and Shakeri, Zahra and Gowda, Vrushab and Samsel, Konrad and Bedayat, Arash and Ghasemiesfe, Ahmadreza and Bagci, Ulas and Hsiao, Albert and Leiner, Tim and Carr, James and others},
  journal={Radiology},
  volume={317},
  number={2},
  pages={e242961},
  year={2025},
  publisher={Radiological Society of North America}
}

@article{hu2025improving,
  title={Improving Generalization of Medical Image Registration Foundation Model},
  author={Hu, Jing and Yu, Kaiwei and Xian, Hongjiang and Hu, Shu and Wang, Xin},
  journal={IJCNN},
  year={2025}
}

@inproceedings{zheng2024contextual,
  title={Contextual Reinforcement Learning for Unsupervised Deformable Multimodal Medical Images Registration},
  author={Zheng, Yang and Xian, Hongjiang and Shuai, Zhikun and Hu, Jing and Wang, Xin and Hu, Shu},
  booktitle={2024 IEEE International Joint Conference on Biometrics (IJCB)},
  pages={1--9},
  year={2024},
  organization={IEEE}
}

@article{tsai2024uu2,
  title={UU-Mamba: Uncertainty-aware U-Mamba for Cardiovascular Segmentation},
  author={Tsai, Ting Yu and Lin, Li and Hu, Shu and Tsao, Connie W and Li, Xin and Chang, Ming-Ching and Zhu, Hongtu and Wang, Xin},
  journal={arXiv preprint arXiv:2409.14305},
  year={2024}
}

@inproceedings{tsai2024uu,
  title={UU-Mamba: uncertainty-aware u-mamba for cardiac image segmentation},
  author={Tsai, Ting Yu and Lin, Li and Hu, Shu and Chang, Ming-Ching and Zhu, Hongtu and Wang, Xin},
  booktitle={2024 IEEE 7th International Conference on Multimedia Information Processing and Retrieval (MIPR)},
  pages={267--273},
  year={2024},
  organization={IEEE}
}

@inproceedings{lin2024robust,
  title={Robust covid-19 detection in ct images with clip},
  author={Lin, Li and Krubha, Yamini Sri and Yang, Zhenhuan and Ren, Cheng and Le, Thuc Duy and Amerini, Irene and Wang, Xin and Hu, Shu},
  booktitle={2024 IEEE 7th International Conference on Multimedia Information Processing and Retrieval (MIPR)},
  pages={586--592},
  year={2024},
  organization={IEEE}
}

@inproceedings{zhu2024cgd,
  title={CGD-Net: A Hybrid End-to-end Network with gating decoding for Liver Tumor Segmentation from CT Images},
  author={Zhu, Xiaogang and Liu, Tao and Liu, Ziqiu and Shaobo, Ouyang and Wang, Xin and Hu, Shu and Ding, Feng},
  booktitle={2024 IEEE International Conference on Advanced Video and Signal Based Surveillance (AVSS)},
  pages={1--7},
  year={2024},
  organization={IEEE}
}

@article{wang2024u,
  title={U-medsam: Uncertainty-aware medsam for medical image segmentation},
  author={Wang, Xin and Liu, Xiaoyu and Huang, Peng and Huang, Pu and Hu, Shu and Zhu, Hongtu},
  journal={arXiv preprint arXiv:2408.08881},
  year={2024}
}

@inproceedings{hu2024umednerf,
  title={UMedNeRF: Uncertainty-aware single view volumetric rendering for medical neural radiance fields},
  author={Hu, Jing and Fan, Qinrui and Hu, Shu and Lyu, Siwei and Wu, Xi and Wang, Xin},
  booktitle={2024 IEEE International Symposium on Biomedical Imaging (ISBI)},
  pages={1--4},
  year={2024},
  organization={IEEE}
}

@article{wang2024neural,
  title={Neural radiance fields in medical imaging: A survey},
  author={Wang, Xin and Chen, Yineng and Hu, Shu and Fan, Heng and Zhu, Hongtu and Li, Xin},
  journal={arXiv preprint arXiv:2402.17797},
  year={2024}
}
}


\end{document}